# Upper Limb Movement Recognition utilising EEG and EMG Signals for Rehabilitative Robotics

Zihao Wang[1] and Ravi Suppiah[2]

[1,2]National University of Singapore, 21 Lower Kent Ridge Rd, Singapore, 119077
`e0424629@u.nus.edu`

**Abstract.** Upper limb movement classification, which maps input signals to the target activities, is a key building block in the control of rehabilitative robotics. Classifiers are trained for the rehabilitative system to comprehend the desires of the patient whose upper limbs do not function properly. Electromyography (EMG) signals and Electroencephalography (EEG) signals are used widely for upper limb movement classification. By analysing the classification results of the real-time EEG and EMG signals, the system can understand the intention of the user and predict the events that one would like to carry out. Accordingly, it will provide external help to the user. However, the noise in the real-time EEG and EMG data collection process contaminates the effectiveness of the data, which undermines classification performance. Moreover, not all patients process strong EMG signals due to muscle damage and neuromuscular disorder. To address these issues, this paper explores different feature extraction techniques and machine learning and deep learning models for EEG and EMG signals classification and proposes a novel decision-level multisensor fusion technique to integrate EEG signals with EMG signals. This system retrieves effective information from both sources to understand and predict the desire of the user, and thus aid. By testing out the proposed technique on a publicly available WAY-EEG-GAL dataset, which contains EEG and EMG signals that were recorded simultaneously, we manage to conclude the feasibility and effectiveness of the novel system.

## 1   Introduction

Due to the fast development of robotic control, artificial robotic technology has unlimited potential in many realms, particularly in rehabilitation. The designs such as exoskeleton and robotic prosthetic designs can help a person overcome injuries and ameliorate the present suffering. Hand prostheses controlled by bioelectrical means, constitute the type of artificial limb with the highest degree of rehabilitation. This is because they can synthesise the aesthetic aspect, the strength and speed of grasping, as well as many possibilities of adaptation to different degrees of disability [1]. For these designs to function properly in upper limb rehabilitation, the upper limb classification technique is of paramount importance as it maps sensor data to the desired events that the users intend to perform. Many prior studies show significant progress in upper limb movement classification. Nazari et al. perform a three-class classification of knee movements (i.e., knee exercises in sitting and standing positions and normal walking) using knee angle data for participants without knee abnormalities and achieve a cross-validation performance accuracy of 84.1% using Support Vector Machine [2]. Biswas et al. propose a methodology to classify three upper limb movements (i.e., extension, flexion and rotation) using accelerometer data and gyroscope data and achieve a 10-fold cross-validation accuracy of 88% using accelerometer data and 83% using gyroscope data for healthy participants, by making use of Linear Discriminant Analysis classifier and Support Vector Machine [3]. Other than inertial sensor data, Electromyography (EMG) and Electroencephalography (EEG) signals are used ubiquitously for upper limb movement classification as well.

Proper interpretation of EMG signals should allow computation of the desired human motion in advance, before the muscles contract and reaction with the mechanical construction takes place [4]. Since there is a correspondence between muscle activity and EMG signals, information can be extracted from them to identify different movements, recognise the users' intention for a specific gesture such as grasping an object or pointing at something and detect the users' intention to perform some tasks. Guo et al. explore the differences in performance among four feature extraction techniques and two classifiers by performing an eight-class classification on upper limb motions using surface EMG signals and achieve a training accuracy of 88.7% using neural networks and 85.9% using Support Vector Machine [5]. Burns et al. propose a methodology for upper limb motor assessments for stroke patients using surface EMG signals and achieve an average classification accuracy of 75.5% using discrete wavelet transform and the enhanced probabilistic neural network [6]. Cene et al. combine two strategies to enhance the representativity of the surface EMG signals. They introduce a stochastic filter based on Antonyan Vardan Transform for data cleaning to reduce the stochastic behaviour of the signals and propose a novel surface EMG feature called Differential Enhanced Signal [7].

Moreover, due to the development of signal processing techniques, recently researchers have explored the possibility of using EEG signals for upper limb classification. Caracillo et al. perform a four-class classification for upper limb movement recognition using EEG signals and achieve a classification accuracy of 49.36% using a Linear Discriminant Analysis classifier [8]. Samuel et al. propose a linear feature combination technique for motor imaginary upper limb movements classification using EEG data and achieve an accuracy of 90.68% using spectral domain descriptors and 99.55% using time domain descriptors [9]. Ofner et al. perform a six-class classification for upper limb movements using low-frequency time-domain information from EEG signals and achieve an average classification accuracy of 55% for real movements and 27% for motor imaginary movements. They also decode the brain areas that contain discriminative movement information [10].

However, building the rehabilitative system with EMG or EEG signals alone restricts the user group as not all types of patients process effective and recognizable EMG and EEG signals, especially for amputees and patients who suffer from neuromuscular disorder.

Furthermore, the presence of noise in the real world undermines the effectiveness and cleanness of EEG and EMG signals. To build a more general rehabilitative system targeting a greater group of users with acceptable performance, we will introduce a novel decision-level multisensor data fusion technique. In particular, the new system will integrate EMG signals with EEG signals. The target audience of our research is those who experience either neurological challenges or physical challenges and need external help for rehabilitative purposes. The objective of this paper is to propose a fusing technique to combine multi-class classification results on EEG signals and EMG signals, which can be applied to a rehabilitative system to help patients aiming for upper limb rehabilitation. To achieve this objective, this research will focus on EEG-based and EMG-based multi-class classification and fuse the results from both sources to provide a more accurate prediction. In the end, based on the experiment results on a publicly available WAY-EEG-GAL dataset, we manage to conclude the effectiveness and feasibility of the proposed multisensor data fusion technique on upper limb classification for rehabilitative robotics.

## 2      Methodology

Studies have been conducted for EEG-EMG data fusion. Hooda et al. perform a five-class classification for unilateral foot movements and achieve a maximum prediction accuracy of 96.58% for a cascaded scheme using the EEG-EMG fusion technique, and they explore the classification performance on different channel sets [11]. Tryon et al. develop convolutional neural network models using combined time-frequency domain EEG-EMG inputs and achieve an average accuracy of 80.51% for a three-class classification [12]. Sbargoud et al. pre-process the EEG and EMG data through wavelet packets transform and train the artificial neural network classifier, followed by a fusion using belief theory [13].

  This paper explores various feature extraction techniques and classifiers for EEG and EMG signals classification, introduces a novel decision-level data fusion algorithm that can divulge information from both EEG and EMG, and validates the performance by introducing Gaussian noise to the dataset, which the prior works fail to conduct. Moreover, we will evaluate the effectiveness of the proposed methodology by using a different dataset as compared to prior works, which is the WAY-EEG-GAL dataset [14], to demonstrate the effectiveness and universality of the EEG-EMG data fusion technique on different EEG-EMG data acquisition techniques and on variety of events.

  In summary, section 3 introduces basic information of the WAY-EEG-GAL dataset, section 4 and 5 investigate EEG and EMG classification respectively, section 6 proposes and verifies a decision level EEG-EMG data fusion technique, section 7 explores channel reduction for power management.

## 3      Introduction to the EEG and EMG dataset

This research uses a publicly available WAY-EEG-GAL dataset to carry out signal analysis [14][15]. The dataset includes EEG and EMG recordings from humans who perform a grasp-and-lift task. During the experimental process, EEG and EMG signals were recorded simultaneously. A total of twelve participants were invited to carry out the whole experimental process. Each participant was required to perform 6 weight series (i.e., 34 lifts per series), 2 surface series (i.e., 34 lifts per series) and 2 mixed series (i.e., 28 lifts per series). During the weight series, the object's weight was changed between 165, 330 and 660g randomly. During the surface series, the object's surface fiction was changed among sandpaper, suede, and silk randomly. Thus, a total of 3,936 trials were carried out. The participant's task in each trial was to reach for a small object, grasp and lift it a few centimetres up, hold it stably for a few seconds, then release the object and withdraw one's hand back to the initial position. During the whole experimental process, EEG signals were recorded in 32 channels with a sampling rate of 500 Hz and EMG signals were recorded in 5 channels with a sampling rate of 4000 Hz. Moreover, the dataset also contains time point information such as the instance when the LED is on and the instance when the LED is off, which can be used to separate different events that can be used in the classification process later. Specifically, from the time point information, we can divide each trial into four different stages to carry out a four-class classification of EEG and EMG signals.

## 4     Electroencephalography
### 4.1     Introduction to Electroencephalography

Electroencephalography (EEG) is a technique to record the electrical activity on the scalp that has been shown to represent the macroscopic activity of the surface layer of the brain underneath. EEG measures voltage fluctuations generated by an ionic current within the neurons of the brain. In short, an EEG records the electrical waves of the brain. To retrieve the EEG signals, multiple electrodes are placed on the scalp (see Fig. 1), and they are named respectively for differentiation. The exact placements of the electrodes are shown in Fig. 2. For example, the electrodes that are placed in the centre of the scalp are named with the prefix 'C' (i.e., C3, C4, Cz etc.).

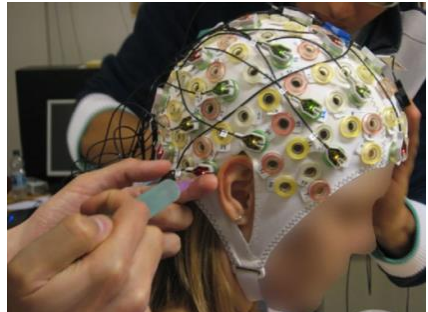 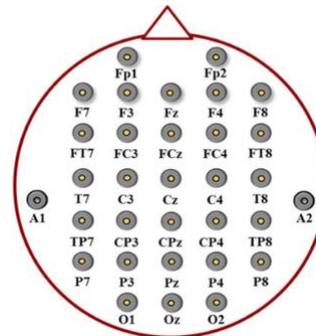

**Fig. 1.** Electrode placements for EEG [14]     **Fig. 2.** Electrode placements for 32-channel EEG [16]

### 4.2     Literature review on EEG signals classification

Prior works have been conducted on EEG signal classification in human activity recognition, which achieves promising results.

Ahmad et al. have done a binary classification on the eyes open and intelligent test (IQ task) [17]. They first perform a Discrete wavelet transform to obtain wavelet coefficients, sample entropy and approximate entropy respectively. The classifier that they choose is a support vector machine. In summary, they obtain 87.5% classification accuracy with linear features and 92.1% classification accuracy with non-linear features.

Chakraborti et al. introduce a methodology for EEG signals classification [18]. They extract data from C3 and C4 channels and band-pass filtered using an elliptical filter (i.e., Order 14) between 8 and 30 Hz, followed by Wavelet transformation to obtain wavelet coefficients and power spectrum estimates using the Welch approach. Lastly, they fit the data to support vector machines with the RBF kernel (RSVM). For left-right hand movement, the maximum classification accuracy of 87.50% is obtained using the wavelet coefficient for the RSVM classifier. Moreover, by using power spectral density (PSD) as the feature set, the maximum classification accuracy of 87.35% is obtained for the RSVM classifier. For the multi-class classification (i.e., Finger-Elbow-Shoulder classification), the maximum classification accuracy of 80.11% for the elbow, 93.26% for the finger and 81.12% for the shoulder are obtained using the features obtained from power spectral density for RSVM classifier. While taking the wavelet coefficient as a feature set, the maximum classification accuracy of 74.24% for the RSVM classifier.

Gomez-Rodriguez et al. have carried out a classification among flexion, extension of the forearm and motor imagery of the forearm (also flexion or extension) [19]. They apply Welch's method to compute power spectral density for 2 Hz frequency bins in the frequency range (2 - 42 Hz) and use the results to fit a support vector machine.

Jeong et al. introduce a methodology for classification using EEG and EMG signals. For EEG signals classification [20], they first apply a zero-phase fourth-order Butterworth filter for band-pass signal filtering to filter between 8 and 30 Hz ($\mu$ and $\beta$ frequency bands) respectively. Next, they apply independent component analysis (ICA) to obtain the correct EEG signal by removing artefacts such as eye and head movement artefacts, followed by applying a common spatial pattern (CSP) algorithm as a feature extraction method. Lastly, they apply a regularised linear discriminant analysis (RLDA) method as the classification method.

Kumarasinghe et al. introduce a Brain-Inspired Spiking Neural Network (BI-SNN) architecture which can learn and reveal deep in time-space functional and structural patterns from spatiotemporal data [21]. Moreover, they introduce an algorithmic methodology to explore the functional and structural interactions of brain networks in different granularity levels using the brain inspired NeuCube SNN [21]. Furthermore, Kumarasinghe et al. introduce a novel neural decoder in 2021, namely Brain-inspired spiking neural networks (BI-SNN) and test its classification performance using the WAY-EEG-GAL dataset [22]. In short, BI-SNN is the combination of eSPANNet and NeuCube. In their research, they use EEG signals as input and EMG signals as expected output. Firstly, they use a spike encoding algorithm for data pre-processing. Secondly, they train BI-SNN using alpha, beta, and gamma frequency bands of EEG signals. Thirdly, they compare the performance of BI-SNN with the Generalised Linear Model (GLM) in four different aspects, namely accuracy, interpretability, prediction latency and training speed. Eventually, they conclude the superiority and efficiency of BI-SNN.

Liu et al. introduce a three-branch 3D convolutional neural network (CNN) to complete feature extraction and classification, which aims to resolve the issues of class imbalance and "easy-hard" examples (i.e., the different classification difficulties for different classes) [23]. Firstly, they do a 3D representation of the EEG signal. Secondly, they divide the data into four different motion stages. The first movement stage is when hands start to move. The second movement stage is when the finger starts to apply load force. The third movement stage is when fingers apply load force and hands return objects. The fourth movement stage is when the participants return the hand to its original position. After dividing the EEG signals into four different stages, they apply cropped strategies to balance different kinds of EEG data and they introduce a focal loss function to address "easy-hard" examples to improve classification accuracy. Finally, they compare classification accuracy among different techniques, namely Filter Bank Common Spatial Pattern (FBCSP), Channel-wise Convolution with Channel Mixing (C2CM) and multi-branch 3-dimensional convolutional neural network (multi-branch 3D CNN).

### 4.3 EEG data visualisation

In this research, we use the EEG recordings of lifting a 330g object. All results that are used in this analysis have the same parameters (i.e., 330g and sandpaper) to maintain consistency and a total of 7 channels are selected out of 32 channels, namely C3, C4, CP1, CP2, CP5, CP6 and Cz channels. These channels are in the central region of the scalp, so they are considered informative for EEG signals analysis, and they are commonly used in much EEG-based research. Through data visualisation of the selected EEG signals in the time domain, we can observe that no obvious pattern can be discovered (see Fig. 3), leading to the necessity of analysis in the frequency domain. Moreover, from data visualisation of the selected EEG signals in the frequency domain, we can observe that low-frequency components (i.e., 0 - 8 Hz) are noisy (see Fig. 4) as low-frequency components contain artefacts such as eye and head movement artefacts which should not be included for analysis. Moreover, the alpha frequency band (i.e., 8 - 12 Hz) and beta frequency band (i.e., 12 - 30 Hz) are the most informative as their amplitudes are relatively higher than the rest, which can be verified from the spectrograms of the selected EEG signals as well (see Fig. 5). More importantly, alpha and beta frequency bands are proven to be effective in EEG-based classification [24].

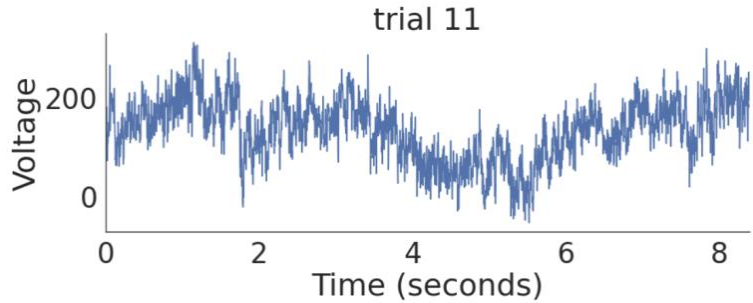

**Fig. 3.** Time domain visualisation for Trail 11 from the C3 channel

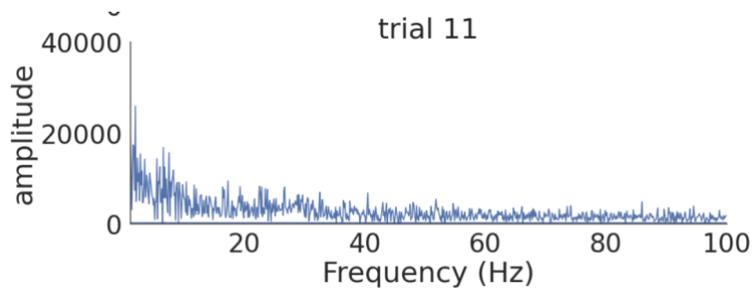

**Fig. 4.** Frequency domain visualisation for Trail 11 from the C3 channel

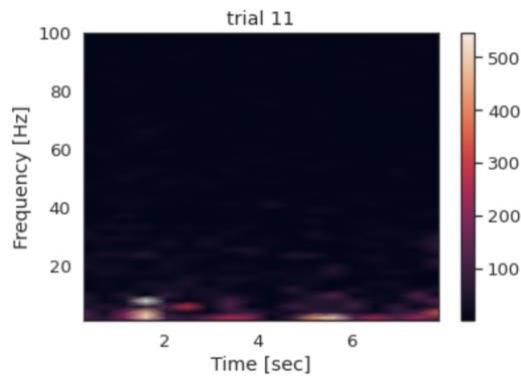

**Fig. 5.** Spectrogram for Trail 11 from the C3 channel

### 4.4  Feature engineering from time and frequency domains for EEG signals

Following the classic time-series data pre-processing techniques, we first extract features from time and frequency domains. Using a fixed sliding window size of 0.1 seconds with overlapping of 0.08 seconds, we have extracted three-time domain features from the selected 7 channels respectively, namely, standard deviation, mean absolute and variance.

*Mean Absolute Value (MAV):*

$$\text{MAV} = \frac{1}{N}\sum_{i=1}^{N}|x_i| \quad (1)$$

*Standard Deviation (SD):*

$$\text{SD} = \sqrt{\frac{1}{N}\sum_{i=1}^{N}(x_i - \overline{x})^2} \quad (2)$$

*Variance (V):*

$$\text{V} = \frac{1}{N}\sum_{i=1}^{N}(x_i - \overline{x})^2 \quad (3)$$

Moreover, using the same sliding window size, we have extracted five frequency domain features after fast Fourier Transform from the selected 7 channels respectively, namely sub-band power of alpha frequency band (i.e., 8 - 12 Hz), sub-band power of beta frequency band (i.e., 12 - 30 Hz), peak power spectrum density (PSD) (i.e., using Welch's method), the frequency with the maximum PSD and spectral energy.

*Absolute Sub-band power (ASB):*
To calculate the absolute sub-band power of EEG signals, we first apply Welch's method to estimate the power spectral density (see Fig. 6). Welch's method is to average consecutive fast Fourier Transform of small windows of the signals and the optimal window duration is defined to be sufficiently long to hold at least two full cycles of the lowest frequency of interest (i.e., $\frac{2}{Lowest\ Frequency\ of\ Interest}$). The highlighted blue area in Fig. 6 is equal to the alpha sub-band power of EEG signals. To approximate the blue area, we will make use of Simpson's rule, which is to decompose the area into several parabolas and sum the area of these parabolas.

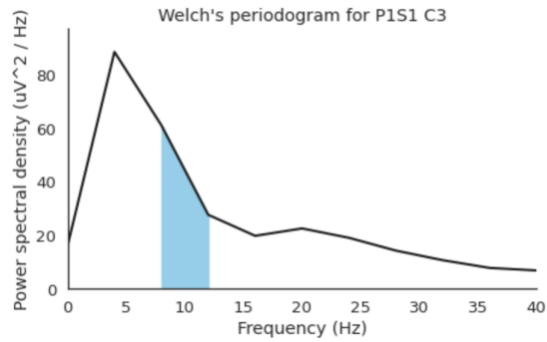

**Fig. 6.** Welch's periodogram

*Peak Power Spectrum Density (PPSD):*
After using Welch's method to approximate the power spectral density (PSD), we remove the low-frequency components (i.e., less than 1 Hz) and retrieve the maximum power spectral density from the remaining frequency band.

$$\text{PSD} = \lim_{T \to \infty} \frac{1}{2T} |X_T(f)|^2 \text{ where } X_T(f) = \int_{-\infty}^{\infty} x(t) \exp(-j2\pi f t) dt \qquad (4)$$

*Frequency with the Peak Power Spectral Density (FPPSD):*
After generating the PPSD, we retrieve the corresponding frequency as one of the features.

*Spectral Energy (SE):*

$$\text{SE} = |X(f)|^2 \text{ where } X(f) = \int_{-\infty}^{\infty} x(t) \exp(-j2\pi f t) dt \qquad (5)$$

Furthermore, using the same sliding window size, we have extracted two time-frequency domain features using Discrete Wavelet Transform from the selected 7 channels respectively, namely approximation coefficients (cA) and detail coefficients (cD). In summary, there are 10 distinct features for 7 selected channels, leading to a total of 70 features. After feature extraction, we apply the Boruta algorithm[1] [25] for feature selection, which makes use of the random forest classifier. As a result, we dispose of 19 features as they do not give a clear differentiation for our classification problem, and we select the remaining 51 features to train the classifiers.

### 4.5   Classification performance for EEG signals

From the description of the dataset, we can divide each trial into four different stages. The first stage is resting, while the whole upper limb remains steady. The second stage is forearm and elbow extension, while the participant moves the arm forward to reach the object and applies no load force on the fingers. The third stage is lifting, while the participant applies load force on the fingers to lift the objects. The fourth stage is forearm and elbow flexion, while the participant moves the arm back to return to the original position and applies no load force on the fingers. Thus, we will apply a four-class classification to the EEG data after features extracted from time and frequency domains.

Next, for time and frequency domain features, we split and scale the data into training and testing sets respectively (i.e., splitting is done before scaling to prevent data snooping) and fit them into five different classifiers, namely multilayer perceptron (MLP), support vector machine with the linear kernel (LSVM), SVM with Radial Basis Function kernel, k-nearest neighbour (kNN) and Long short-term memory (LSTM), to evaluate their performance respectively. The kNN classifier is used as the baseline model to compare the performance of various classifiers and SVM with kernel functions is commonly used in the prior works, which is considered state-of-the-art. MLP is chosen as the basic feed-forward artificial neural network to learn a nonlinear relationship. LSTM is chosen as the recurrent neural network, which is proven to be effective on time-series data classification tasks. The classification reports (i.e., accuracy, precision, recall and F1-score) for these five classifiers are obtained respectively by using 10-fold cross-validation. By considering the runtime, kNN is not suitable for real-time classification as the runtime for kNN is O(n), where n is the number of training data points. From the performance summary (see Table 1), we can observe that LSTM will be most suitable for the real-time multi-class classification of EEG data.

---

[1] Boruta algorithm is a feature selection algorithm that selects features down to the "all-relevant" stopping points, instead of the "minimal-optimal" stopping points.

|  | Time and frequency domain features | | | |
|---|---|---|---|---|
|  | Accuracy | Precision | Recall | F1-score |
| Multilayer perceptron | 0.495 | 0.493 | 0.493 | 0.531 |
| SVM with linear kernel | 0.559 | 0.559 | 0.559 | 0.559 |
| SVM with RBF kernel | 0.579 | 0.579 | 0.579 | 0.579 |
| K-nearest Neighbour | 0.477 | 0.477 | 0.477 | 0.477 |
| LSTM | 0.827 | 0.827 | 0.827 | 0.827 |

**Table 1.** 10-fold cross-validation performance summary for EEG

## 5     Electromyography
### 5.1     Introduction to Electromyography

Electromyography (EMG) is a technique to record the electrical activity produced by skeletal muscles. EMG measures electric potential resulting from muscle cells when they are neurologically activated. In short, an EMG evaluates nerve and muscle function in the arms or legs. To retrieve the EMG signals, multiple electrodes are placed on the muscles (see Fig. 7). For the WAY-EEG-GAL dataset, electrodes are placed on five different muscles, namely Anterior Deltoid, Brachioradialis, Flexor Digitorum Profundis, Common Extensor Digitorum, and First Dorsal Interosseous. Anterior Deltoid is the front delts that help to move one's arm forward. Brachioradialis is the muscle in the lower part of the arm that helps the arm bent at the elbow. Flexor Digitorum Profundis is the muscle in the forearm of humans that flexes the fingers. Common Extensor Digitorum is the muscle of the posterior forearm that extends the phalanges, then the wrist, and finally the elbow. It tends to separate the fingers as it extends them. First Dorsal Interosseous is the muscle on the radial side of the second metacarpal and the proximal half of the ulnar side of the first metacarpal which abducts the thumb and the index finger.

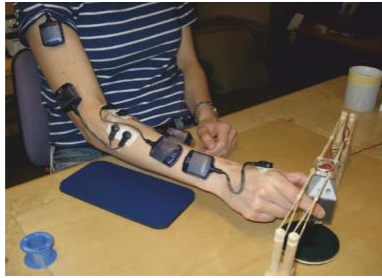
**Fig. 7.** Demonstration of electrode placements for EMG [14]

### 5.2 Literature review on EMG signals classification

Prior works have been done on EMG signal classification in human activity recognition, which achieves promising results.

Abbaspour et al. propose a new feature set for classification from the time domain, frequency domain (i.e., after fast Fourier Transform) and time-frequency domain (i.e., after Discrete Wavelet Transform), followed by Principal Component Analysis to reduce the dimensionality of the feature set [26]. Six classifiers are used, which are LDA, k-nearest neighbour, decision tree, maximum likelihood estimation, SVM and multilayer perceptron.

Batzianoulis et al. have decoded the grasp intention during the reach-to-grasp motion [27]. The event is divided into three phases and the goal of this research is to differentiate five different grasp types through the analysis of the three phases during the reach-to-grasp motion. The first phase is when the angular velocity of the elbow joint exceeds a velocity threshold (i.e., the velocity of the motion increases). The second phase is when the angular velocity of the elbow drops below a velocity threshold (i.e., the velocity of the motion decreases). The third phase is after the completion of the elbow extension (i.e., reaching motion is completed). Four classifiers are used, which are an LDA classifier, an SVM with linear kernel, an SVM with a Radial Basis Function kernel and Echo state networks (i.e., a recurrent neural network RNN). For LDA and SVM, three features were chosen, which are the average activation of each time window, its waveform length, and the number of slope changes.

Gokgoz et al. investigate the classification performance of different versions of decision tree algorithms (i.e., classification and regression tree, C4.5 and random forest) with Discrete Wavelet Transform coefficients using intramuscular EMG [28].

Ortiz-Catalan et al. introduce a novel platform for the control of artificial limbs [29]. The platform is implemented as a collection of functions and GUIs divided into the following modules: signal recordings, signal treatment, signal features, pattern recognition, and control. For signal treatment, it has been shown through information theory that EMG windows of 100 to 300 ms contain the highest information content. For signal features, mean absolute value, zero crossing, slope sign changes, and waveform length are used as the feature set. For pattern recognition, regulatory feedback networks, linear discriminant analysis and multilayer perceptron are used as classifiers.

Phinyomark et al. conclude the optimal representative EMG feature sets, which are mean absolute values, waveform length, Willison amplitude, 4th order auto-regressive coefficients and mean absolute value slope [30]. They also emphasise that frequency domain features are not effective for EMG classification tasks.

Simao et al. provide a review on decoding for surface EMG and intramuscular EMG [31]. They conclude that mean absolute value, waveform length, Willison amplitude and auto-regressive coefficients are the most effective features for EMG decoding and time domain features perform better than frequency domain features. PCA, non-negative matrix factorization and independent component analysis are recommended for dimensionality reduction and one-class SVM, the single class minimax probability machine, kernel PCA and LDA are recommended as classifiers.

### 5.3    EMG data visualisation

The EEG signals and EMG signals are recorded simultaneously and the trials we use are identical for EEG and EMG signals classification. Like EEG signals, in this research, we use the EMG recordings of lifting a 330g object. A total of 5 channels are used, which are Anterior Deltoid, Brachioradialis, Flexor Digitorum Profundis, Common Extensor Digitorum, and First Dorsal Interosseous. From the data visualisation in the time domain, we can easily differentiate different events, especially between the resting stage and the moving stage (see Fig. 8). From the data visualisation in the frequency domain, we can observe that frequency domain features are less informative compared to time domain features (see Fig. 9), which is emphasised by [30] (i.e., Frequency domain features are not effective for EMG signals classification). From the spectrogram, we can observe that different muscles are triggered at different time steps to perform different events (see Fig. 10).

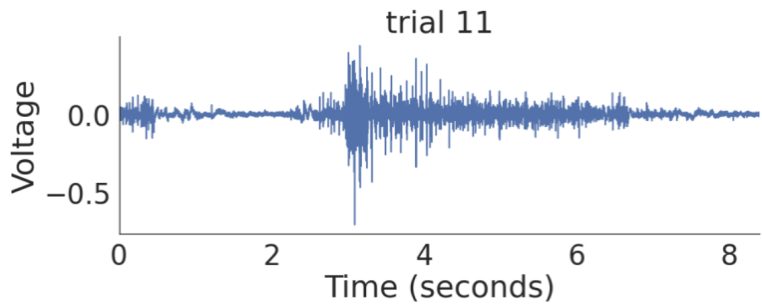

**Fig. 8.** Time domain visualisation for Trail 11 from the First Dorsal Interosseous channel

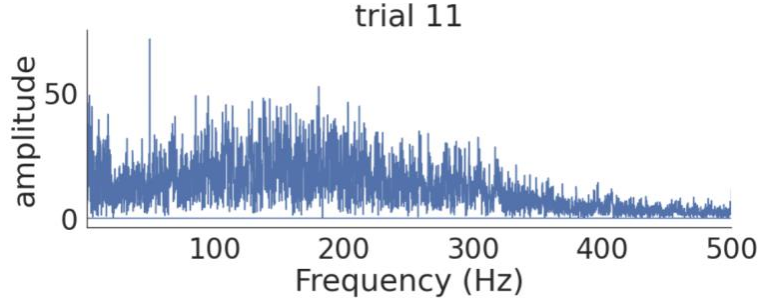

**Fig. 9.** Frequency domain visualisation for Trail 11 from the First Dorsal Interosseous channel

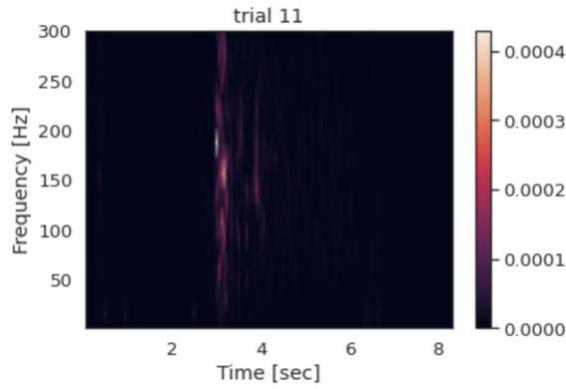

**Fig. 10.** Spectrogram for Trail 11 from First Dorsal Interosseous channel

### 5.4  Feature engineering from the time domain for EMG signals

Since frequency domain features are proven to be less effective for EMG signals classification, we extract features from the time domain only. Using a fixed sliding window size of 0.2 seconds with overlapping of 0.15 seconds, we have extracted five-time domain features from the selected 5 channels respectively, namely mean absolute, waveform length, Willison amplitude, mean absolute slope and autoregressive coefficient.

*Mean Absolute Value (MAV):*
$$\text{MAV} = \frac{1}{N}\sum_{i=1}^{N}|x_i| \tag{6}$$

*Waveform Length (WL):*
$$\text{WL} = \sum_{i=1}^{N-1}|x_{i+1} - x_i| \tag{7}$$

*Willison Amplitude (WA):*
$$\text{WA} = \sum_{i=1}^{N-1}[f(|x_n - x_{n+1}|)]$$

$$\text{where } f(x) = 1 \text{ if } x \geq threshold, f(x) \; 0 \text{ if otherwise} \qquad (8)$$

*Mean Absolute Slope (MAS):*
$$MAS_k = MAV_{k+1} - MAV_k \text{ where } k = 1, 2, 3, \ldots, K - 1 \qquad (9)$$

*Autoregressive coefficients:*
Use the original data to fit and train the autoregressive (AR) model and obtain the coefficients $a_p$ as a feature vector. AR model is a time series model that uses observations from previous time steps as input to a regression equation to predict the value at the next time step. It can be expressed as

$$x_i = \sum_{p=1}^{P} a_p x_{i-p} + w_i$$
$$\text{where } P \text{ is the order of the AR model and } w_i \text{ is the bias term} \qquad (10)$$

In summary, there are 5 distinct features for 5 selected channels, leading to a total of 25 features. After feature extraction, we apply the Boruta algorithm[2] [25] for feature selection, which makes use of the random forest classifier. As a result, all 25 features give a clear differentiation for our classification problem and thus we select all 25 features to train the classifiers.

### 5.5   Classification performance for EMG signals
From the description of the dataset, like EEG, we can divide each trial into four different stages, namely resting, forearm and elbow extension, lifting and forearm and elbow flexion. Thus, we will apply a four-class classification to the EMG data after features extracted from the time domain.

Next, similar to EEG signals training and testing, for the time domain feature, we split and scale the data into training and testing sets respectively and fit them into five different classifiers, namely multilayer perceptron (MLP), support vector machine with the linear kernel (LSVM), SVM with Radial Basis Function kernel, k-nearest neighbour (kNN) and Long short-term memory (LSTM), to evaluate their performance respectively. The classification reports (i.e., accuracy, precision, recall and F1-score) for these five classifiers are obtained respectively by using 10-fold cross-validation. From the performance summary (see Table 2), we can observe that all classifiers yield high accuracy and LSTM will be most suitable for real-time multi-class classification on EMG data.

---

[2] Boruta algorithm is a feature selection algorithm that selects features down to the "all-relevant" stopping points, instead of the "minimal-optimal" stopping points.

|  | Time and frequency domain features | | | |
| --- | --- | --- | --- | --- |
|  | Accuracy | Precision | Recall | F1-score |
| Multilayer perceptron | 0.993 | 0.992 | 0.994 | 0.994 |
| SVM with linear kernel | 0.997 | 0.997 | 0.997 | 0.997 |
| SVM with RBF kernel | 0.990 | 0.990 | 0.990 | 0.990 |
| K-nearest Neighbour | 0.987 | 0.987 | 0.987 | 0.987 |
| LSTM | 0.998 | 0.998 | 0.998 | 0.998 |

**Table 2.** 10-fold cross-validation performance summary for EMG

# 6 Fusion of EEG and EMG
## 6.1 Classification results comparison

In general, the multi-class classification results for EMG signals are relatively higher compared to those of EEG signals for the same classifier. Moreover, we can notice that for both signals, LSTM is the most effective classifier as it yields the highest accuracy. EEG signal classification is much more difficult compared to EMG signal classification due to the indirect connection between brain signals and the desired events. However, with the choice of an appropriate feature extraction technique and classifier, it still yields a relatively high accuracy of around 80%. Hence, it is feasible to integrate EEG signals with EMG signals for the control of rehabilitative robotics.

## 6.2 Proposed fusion algorithm

*Weightage of EEG classification result ($W_{eeg}$):*

$$W_{eeg} = \frac{E_{eeg}}{E_{eeg} + E_{emg}} = \frac{A_{eeg}}{A_{eeg} + A_{emg}} \tag{11}$$

where $E_{eeg}$ is the expected prediction accuracy of EEG, $E_{emg}$ is the expected prediction accuracy of EMG, $A_{eeg}$ is the 10-fold cross-validation accuracy of EEG, and $A_{emg}$ is the 10-fold cross-validation accuracy of EMG.

*Weightage of EMG classification result ($W_{emg}$):*

$$W_{emg} = \frac{E_{emg}}{E_{eeg} + E_{emg}} = \frac{A_{emg}}{A_{aeg} + A_{emg}} \tag{12}$$

where $E_{eeg}$ is the expected prediction accuracy of EEG and $E_{emg}$ is the expected prediction accuracy of EMG, $A_{eeg}$ is the 10-fold cross-validation accuracy of EEG, and $A_{emg}$ is the 10-fold cross-validation accuracy of EMG.

*Degree of noisiness (N):*

$$N = \frac{\frac{1}{o}\sum_{k=1}^{o}|x'_{k+1} - x'_k|}{\frac{1}{m}\sum_{i=1}^{m}\frac{1}{n^{[i]}}\sum_{j=1}^{n^{[i]}}|x_{j+1}^{[i]} - x_j^{[i]}|}$$  (13)

where $x'$ is the unseen trail of data, $x$ is the trails of training data, $o$ is the length of the unseen trail of data, $n$ is the length of the $i^{th}$ trial of training data, and $m$ is the total number of training trails. The numerator computes the average fluctuation across all unseen data. The denominator computes the average fluctuation across all training trials, which is the baseline measure of the noisiness of the data and can be pre-computed. Thus, $N$ measures degree of noisiness of the unseen data.

*Final prediction (F):*

$$F = max(W_{eeg} \times \frac{1}{N_{eeg}} \times Result_{eeg}, W_{emg} \times \frac{1}{N_{emg}} \times Result_{emg})$$  (14)

where $Result_{eeg}$ is the result of EEG classification, $Result_{emg}$ is the result of EMG classification. Final prediction takes either $Result_{eeg}$ or $Result_{emg}$, which depends on degree of truthiness of the data. $\frac{W}{N}$ measures degree of truthiness of the classification result. Since the expected performance of the EEG classifier and the expected performance of the EMG classifier is different, the variable $W$ is introduced to weigh the performance of the two classifiers. Moreover, the variable $N$ is introduced to measure degree of noisiness of the data. Thus, when degree of noisiness increases, degree of truthiness decreases.

### 6.3    Verification of the proposed fusion algorithm

To verify the proposed EEG/EMG data fusion technique, we introduce a moderate level of Gaussian noise to EEG and EMG data and input it to the trained LSTM classifiers and the fusion algorithm. For the first case, we introduce a Gaussian noise to EEG data. Using only the EEG classifier, the accuracy is 65.3%. While using the fusion algorithm, the accuracy is 99.8%. For the second case, we introduce a Gaussian noise to EMG data. Using only the EMG classifier, the accuracy is 89.7%. While using the fusion algorithm, the accuracy is 89.7%. However, when the degree of noisiness increases significantly, the accuracy of the fusion algorithm will be higher than that of the EMG classifier. For the third case, we introduce a Gaussian noise to both EEG and EMG data, the accuracy of the fused algorithm is 89.7%.

In summary, the fusion algorithm retrieves effective information from both EEG and EMG signals. When EEG data is contaminated, it prioritises the results from the EMG classifier. When EMG data is contaminated lightly, it still prioritises the results from the EMG classifier as it has a higher weightage. When EMG data is contaminated significantly, it prioritises the results from the EEG classifier. This fusion algorithm assumes that the input training data is clean.

# 7   Channel reduction for power management

In the case of EMG signals, the electrode placements depend on the muscles that are required to carry out specific events. Thus, the number of channels used for EMG signals depends on the nature of the events that we want to classify. However, in the case of EEG signals, the electrode placements do not depend on the muscles that will be triggered to carry out certain events. Therefore, we can try to reduce the number of channels used for EEG signals for power management purposes by applying ERD/ERS analysis to the EEG signals. Using Event-related desynchronisation (ERD) and event-related synchronisation (ERS) is a relatively novel way for EEG signal decoding. ERD will be detected when one has the desire to move as the brain must desynchronise some neurons to perform the task, while ERS will be detected when one has the desire to rest as the neurons that were previously desynchronised will be synchronised again. Pfurtscheller et al. proposed a band power method for the quantification of ERD/ERS in the time domain in 1999. In this method, a decrease in band power indicates ERD and an increase in band power indicates ERS [32]. In summary, the band power method includes five steps in total [32]. Firstly, apply a bandpass filter to all trials to extract useful information in the targeted frequency band. Secondly, squaring all samples to obtain power samples. Thirdly, averaging all power samples across all trials. Fourthly, smoothing the data and reducing the variability. Lastly, applying baseline correction for better visualisation.

  In this research, we firstly apply a fifth-order Butterworth band pass filter on the original EEG signals to extract information from alpha (i.e., 8 - 12 Hz) and beta (i.e., 12 - 30 Hz) frequency bands for all seven selected channels (i.e., C3, C4, Cz, CP1, CP2, CP5 and CP6) respectively, followed by squaring the results. Then, we crop the data of all trials to maintain the same length and average the cropped data across all trials. Next, we apply a Savitzky-Golay filter for data smoothing. Lastly, we apply baseline correction to the previous results, which takes the period of 1 - 2 seconds as the baseline. From the ERS/ERS result from the C3 channel, we can observe a clear ERD (i.e., a decrease in voltage) at 2 seconds and a clear ERS (i.e., an increase in voltage) at around 7.5 seconds on the beta frequency band (see Fig. 11). However, this pattern cannot be observed from other channels. Thus, we carry out the 4-class classification on EEG signals using the C3 channel only. Unfortunately, the accuracy drops significantly from 0.827 to 0.485 using LSTM. In conclusion, the minimum number of channels used for an effective EEG signal multi-class classification is 7 (i.e., C3, C4, Cz, CP1, CP2, CP5, CP6).

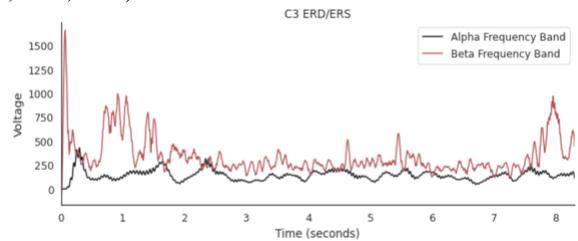

**Fig. 11.** ERD/ERS result from the C3 channel

# 8 Future works and conclusion

EMG signals have different signatures depending on age, muscle development, motor unit paths, skin-fat layer, and gesture [33]. Moreover, EMG signals depend very much on the subject, on skin properties such as skin moisture, as well as on muscle fatigue [4]. Thus, the external appearances of two peoples' gestures might look identical, but the characteristic EMG signals are different. In this study, we implement the system based on inter-participant results. To resolve this issue, in the future we will investigate on data fusion involving more sources such as Electrocardiography (ECG) and Photoplethysmography (PPG) signals.

In conclusion, due to the imperfections in using EEG and EMG signals solely on upper limb movement classification, the classification performance does not achieve its full potential. To address the disadvantages, in this paper, we investigate various feature extraction techniques on state-of-the-art machine learning and deep learning classifiers for EEG and EMG data respectively, introduce a decision-level fusion technique for EEG-based and EMG-based multiclass classification, and verify performance of the proposed LSTM-based EEG-EMG decision level fusion using the WAY-EEG-GAL dataset. Based on the experiment performance, LSTM yields the highest 10-fold cross-validation results on clean EEG (i.e., 82.7%) and EMG (i.e., 99.8%) signals respectively. Unfortunately, after introducing a Gaussian noise to the data, the accuracy for LSTM-based EEG classification declines significantly to 65.3%. With the proposed decision-level fusion, we manage to maintain an average classification accuracy around 89.7% with contaminated EEG and/or EMG signals.

## Data availability